# A Survey on Awesome Korean NLP Datasets

Byunghyun Ban
https://needleworm.github.io
Andong, Republic of Korea
bhban@kakao.com

*Abstract*—English based datasets are commonly available from Kaggle, GitHub, or recently published papers. Although benchmark tests with English datasets are sufficient to show off the performances of new models and methods, still a researcher need to train and validate the models on Korean based datasets to produce a technology or product, suitable for Korean processing. This paper introduces 15 popular Korean based NLP datasets with summarized details such as volume, license, repositories, and other research results inspired by the datasets. Also, I provide high-resolution instructions with sample or statistics of datasets. The main characteristics of datasets are presented on a single table to provide a rapid summarization of datasets for researchers.

## I. Introduction

Natural language processing, NLP, with various neural network models has recently received a large amount of interest from both linguists and AI engineers. NLP has benefited from LSTM [1-3] and RNN models, due to their high performances on sequential data.

Although a CNN has hierarchical structures, YN Dauphin, et al. showed that application of CNN on natural languages outperformed traditional recurrent models [4]. Recent approaches on natural language processing utilize transformer architectures for language processing [5-8].

Noble approaches and brand-new applications inspired the researchers all around the world, to reproduce such high-performance models trained with various languages. For example, H Zhuang, et al. [9] and Y Cui, et al. [10] introduced Chinese based model, and M Al-Smadi, et al. [11] and S Romeo, et al. [12] showed models for Arabic natural language. Also, various Korean based NLP models have been introduced on international conferences [13-16] and Journals [17-18].

Researchers, who are working with non-English NLP technologies or multilanguage models, are highly interested in datasets in different languages. Even for researchers whose native language is a low-share language, the craving for datasets in their mother tongue langue would be much greater. The same goes for the industrial demands. NLP-related product holders are trying to research and develop models in official language of their country. Therefore, natural language processing researchers in Korea are very interested in Korean based NLP datasets. Not only academic institutions [19-20, 25] and individual researchers [21-22], but also large corporations [14, 23-24] are actively producing Korean based natural language datasets.

## II. Related Works

Simon Ruffieux, et al. provided summarized details of human pose estimation datasets to introduce various useful resources to researchers [49]. They wanted to reduce the time which other researchers spend for exploring useful datasets, and to provide useful tips for dataset construction.

Although there are several surveys on NLP technologies and algorithm themselves [26-28] or survey on Korean language processing technologies [29], an academic paper or comprehensive survey on Korean-based NLP dataset is not so many.

Cho, Won Ik, et al. established Open Korean Corpora [40] to provide assorted Korean NLP datasets. They provide the summarized detail of 29 different datasets from text parsing and tagging area, sentence similarity, sentence classification and QA, parallel corpora, multilingual corpora, and speech corpora. They summarized the information of each dataset into a short paragraph of one or two sentences. Their work is suitable for rapid and brief exploration among various datasets. But their work is insufficient to present a rationale for selecting a dataset suitable for the research purpose.

Yeongsook Song published a GitHub Repository named "AwesomeKorean_data" to provide information on various Korean datasets [30]. She introduced 31 datasets with academic references, and 9 datasets by individual researchers, and 5 government driven datasets. AwesomeKorean_data provides summarized detailed of datasets in a table which is easy to read quickly. It also provides hyperlink to the original webpage of each dataset. However, it categorized the license of datasets into too simple format (commercially available or not) so a researcher should figure out additional information on restrictions related to indication of same origin and impossibility of reprocessing and manipulations. Also, it does not provide published applications of datasets.

## III. Survey

I selected 15 datasets which are famous among Korean research communities. Summarization over whole datasets are presented on Table 1. The table shows the name, category, domain and task, size, license, author, year, citation score and related research articles. This section also provides details of each dataset. For example, additional information such as repository URL, format, statistics of dataset or data sample, and application of the data.

TABLE II
DATASET SUMMARIZATION

| Name | Category | Domain & Task | Size | License | Author | Year | Citation Score[e] | Related Research |
|---|---|---|---|---|---|---|---|---|
| Chatbot_data | Conversation | Chatbot conversation (text) | 11.8k | MIT | Song, Youngsook [31] | 2018 | 3 | Yun., et al. (2021) [39] |
| ClovaCall | | Call based dialogue (speech) | 125h, 81k samples | NC/AC[d] | Ha, Jung-woo, et al. [23] | 2020 | 6 | Cho., et al. (2020) [40] |
| KorQuAD | Question Answering | Q&A pairs from Wikipedia articles | 103k | CC BY-ND 2.0 KR | Lim, Seungyoung, et al. [32] | 2019 | 31 | Lee., et al. (2020) [41] |
| Song-NER | Named Entity Recognition | | 30k | CC BY-SA 2.0 | Park, Hyewoong, et al. [33] | 2017 | - | Kim., et al. (2019) [42] |
| KMOUNLP-NER | | Korean word based NER | 24k | NC/AC | Cheon, Minah [20] | 2021 | 5 | Yoo., et al. (2020) [43] |
| Sci-News-Sum-Kr-50 | Text Summarization | News article summarization | 50 | NC/AC | Seol, Jinsuk [34] | 2016 | 3 | Cho., et al. (2020) [40] |
| SAE4K | | Structured Argument Extraction | 50.8k | CC BY-SA 4.0 | Cho, Won Ik, et al. [35] | 2019 | 3 | Cho., et al. (2020) [36] |
| KLUE | NLU[a] | Korean NLU benchmark | 239k | CC BY-SA 4.0 | Park, Sungjoon, et al. [24] | 2021 | 4 | Kim., et al. (2021) [44] |
| KorNLU | NLI[b] / STS[c] | Korean NLI & NLU | NLI: 950k NLU: 8.6k | CC BY-SA 4.0 | Ham, Jiyeon, et al. [14] | 2020 | 9 | Lee., et al. (2020) [45] |
| ParaKQC | STS | Parallel questions and commands | 545k utterances | CC BY-SA 4.0 | Cho, Won Ik, et al. [36] | 2020 | 4 | Park., et al. (2021) [24] |
| NSMC | | Movie reviews & positive/negative indicate | 200k | CC0 1.0 | Park, Eunjeong [37] | 2015 | 50+ | Lee., et al. (2020) [46] |
| Toxic Comment | Semantic Textual Analysis | More specified emotional labels on NSMC | 20k | MIT | Song, Youngsook [21] | 2018 | - | - |
| KHateSpeech | | Malicious comments from news articles | Labeled:9.4k Raw: 2m | CC BY-SA 4.0 | Moon, Jihyung, et al. [38] | 2020 | 13 | Lee., et al. (2021) [47] |
| 3i4k | Semantic Speech Analysis | Utterance intent inference | Corpus:19k Text: 61k Speech: 7k | CC BY-SA 4.0 | Cho, Won Ik, et al. [25] | 2018 | 8 | Cho., et al. (2019) [48] |
| KAIST Corpus | NLU and Machine Translation | Various corpus and Machine Translation | 35 different datasets | NC/AC | Ku-sun Choi [72] | 1997 ~ 2005 | 11 | |

[a] Natural language understanding  [b] Natural language inferences  [c] Semantic textual similarity  [d] Noncommercial and academic use only
[e] Number of citations on 2021.10.15.

## A. Chatbot_data

TABLE II. CHATBOT_DATA OVERVIEW

| Dataset overview | |
|---|---|
| Category | Conversation (text) |
| Size | 11,876 Q&A pairs |
| Format | CSV |
| License | MIT |
| URL | https://github.com/songys/Chatbot_data |
| Author | Song, Youngsook |

| Citation text |
|---|
| Song, Youngsook, Chatbot_data. 2018. Available at https://github.com/songys/Chatbot_data. |

TABLE III. CHATBOT_DATA SAMPLE

| Q&A data pair sample | Label |
|---|---|
| Q1. 12 시 땡!<br>A1. 하루가 또 가네요. | 0 |
| Q2. 3 박 4 일 정도 놀러가고 싶다<br>A2. 여행은 언제나 좋죠 | 0 |
| Q3. PPL 심하네<br>A3. 눈살이 찌푸려지죠. | 0 |

Chatbot_data was published by Youngsook Song in 2018 [31]. The answers were composed of sentences that could comfort a person who has gone through a breakup with a lover. Answer sentences are labeled into 3 classes; 0 for daily life dialogue, 1 for negative (breakup), and 2 for positive (love).

This dataset has MIT license, which allows commercial use, modification, distribution and private use. However, the author is not responsible for any liability or warranty.

Hyeonseo Yun has published a paper on transformer-based AI model to detect unethical sentences from given text data, with this dataset [39].

## B. ClovaCall

TABLE IV. CLOVACALL OVERVIEW

| Dataset overview | |
|---|---|
| Category | Conversation (Speech) |
| Size | 81,222 samples<br>125 hours of raw data<br>67 hours of clean data |
| Format | WAV & JSON |
| License | Noncommercial use for AI Research only |
| URL | https://github.com/clovaai/ClovaCall |
| Author | Jung-Woo Ha, et al. |

| Citation text |
|---|
| Ha, Jung-Woo, et al. "ClovaCall: Korean goal-oriented dialog speech corpus for automatic speech recognition of contact centers." arXiv preprint arXiv:2004.09367 (2020). |

TABLE V. CLOVACALL SAMPLE

**ClovaCall data sample**

```
[
  {
    "wav": "42_0603_748_0_03319_00.wav",
    "text": "단체 할인이 가능한 시간대가 따로 있나요?",
    "speaker_id" : "03319"
  },
  …,
  {
    "wav": "42_0610_778_0_03607_01.wav",
    "text": "애기들이 놀만한 놀이방이 따로 있나요?",
    "speaker_id": "03607"
  }
]
```

ClovaCall was provided by Jung-Woo Ha, et al. in 2020 [23]. It is a large scale Korean speech corpus dataset gathered from recorded phone calls, from more than 11,000 people under a given situation: restaurant reservation. License-free ASR related datasets are old fashioned and most of them are established with English. Therefore, the authors constructed a license-free Korean phone call dataset with baseline codes.

Won Ik Cho, et al. included ClovaCall dataset in their work on Open Korean Corpora [40]. Jihwan Bang, et al. followed a variant version of LAS model proposed on ClovaCall paper [50]. As ClovaCall is recently released, a research article using ClovaCall for training or model validation could not be found yet. However, many recent works have mentioned ClovaCall as related recent works [51-54].

## C. KorQuAD

TABLE VI. KORQUAD OVERVIEW

| Dataset overview | |
|---|---|
| Category | Question Answering |
| Size | 102,960 Pairs |
| Format | JSON |
| License | CC BY-ND 2.0 KR |
| URL | https://korquad.github.io/ |
| Author | Seungyoung Lim, et al. |

| Citation text |
|---|
| Lim, Seungyoung, Myungji Kim, and Jooyoul Lee. "KorQuAD1.0: Korean QA dataset for machine reading comprehension." arXiv preprint arXiv:1909.07005 (2019). |

KorQuAD dataset was published by Seungyoung Lim in 2019 [32]. This dataset is inspired by SQuAD [55], a reading comprehension based question answering dataset driven from Wikipedia articles. It is the largest Korean question answering dataset. The authors drove data from Wikipedia.

TABLE VII. KORQUAD SAMPLE

| KorQuAD Data Sample | Class |
|---|---|
| Q. 외국인들을 위해 먹는 샘물이 일시 판매되었던 년도는 언제일까?<br>A. (전략) 1988년 서울 올림픽 무렵 외국인들을 위하여 일시 판매를 허용했던 적이 있으나, 다시 판매를 제한하였다. (후략) | Short Answer/ Syntax Variation |
| Q. 2009년 시즌 도중 경질된 지바 롯데의 감독은?<br>A. (전략) 시즌 도중에 바비 밸런타인 감독의 해임이 발표되자 일부 팬들은 (후략) | Short Answer/ Vocabulary Variation |
| Q. 피터슨과 노먼 그란츠의 관계는 어떤 과정을 통해 형성되었는가?<br>Title. 오스카 피터슨 – #생애 – #노먼 그란츠 | Long Answer/ Subheading Duplication |
| Q. 이경직의 가족 관계는 어떻게 이루어져 있는가?<br>Title. 이경직 – #가계 | Long Answer/ Subheading Variation |

KorQuAD is actively applied on Korean NLP conferences. Dongheon Lee, et al. applied KorQuAD for benchmark test of reading comprehension model in 2019 [56]. Minho Kim, et al. has developed a machine reading comprehension-based question and answering system and applied KorQuAD for performance test in 2020 [57]. Seohyung Jeon, et al. applied KorQuAD dataset on BERT-based multilingual model for machine reading comprehension task in 2020 [58].

### D. Song-NER

TABLE VIII. SONG-NER OVERVIEW

| Dataset overview | |
|---|---|
| Category | Named Entity Recognition |
| Size | DT 272    LG 29,799<br>PS 11,354    TI 48 |
| Format | TSV |
| License | CC BY-SA 2.0 |
| URL | https://github.com/songys/entity |
| Author | Park, Hyewoong and Song, Youngsook |

**Citation text**

박혜웅, and 송영숙. "음절 기반의 CNN 을 이용한 개체명 인식." 한국어정보학회 학술대회 (2017): 330–332.

Song_NER dataset was established by Hyewoong Park and Yeongsook Song in 2017 [33]. It consists of 4 categorical corpus; date(DT), location(LG), personal name(PS), time(TI). This dataset could be applied for named entity recognition model for chatbot system.

TABLE IX. SONG-NER SAMPLE

| Corpus samples | Category |
|---|---|
| 가윗날, 가을, 가을밤, 가을봄, 가을철, 간밤, 개동 | DT |
| 고양, 광명, 광주, 구리, 구미, 군산, 군포, 김포 | LC |
| 가규, 가도, 가밀, 가사도, 가의, 가중명, 가충 | PS |
| 간밤, 그날밤, 그밤, 긴밤, 날낮, 날밤, 낮, 대낮 | TI |

A chatbot model trained with Song_NER could perform information extraction from user's message. For example, a sentence "내일 저녁 7 시에 이태원 파스타집 5 명 예약해줘 (Please make a reservation for 5 people at the pasta restaurant in Itaewon tomorrow at 7pm)" could processed as below;

> DT : tomorrow
> LC : Itaewon
> PS : 5 people
> TI : 7pm.

Hyunjoong Kim, et al. has included this dataset on Korpora in 2019 [42].

### E. KMOUNLP-NER

TABLE X. KMOUNLP-NER OVERVIEW

| Dataset overview | |
|---|---|
| Category | Named Entity Recognition |
| Size | 23,964 corpuses |
| Format | TEXT |
| License | Noncommercial Research Only |
| URL | https://github.com/kmounlp/NER |
| Author | Cheon, Minah |

**Citation text**

천민아, 다중 생성 단위의 관계 점수를 이용한 학습 말뭉치 생성: 개체명 말뭉치를 중심으로, 한국해양대교 대학원 컴퓨터공학과, 박사학위논문: 2021 년 2 월.

TABLE XI. KMOUNLP-NER SAMPLE

**Tokenized sentence sample**

## 사라예보여인 이스라엘 정착

## 〈사라예보: LOC〉여인 〈이스라엘: ORG〉 정착

| 사라예보 | 사라예보 | NNP | B-LOC |
|---|---|---|---|
| 여인 | 여인 | NNG | O |
| – | – | – | O |
| 이스라엘 | 이스라엘 | NNP | B-ORG |
| – | – | – | O |
| 정착 | 정착 | NNG | O |

KMOUNLP-NER was openly provided at GitHub since 2016, and was officially published by Minah Cheon in 2021 [20]. Her paper proposed a criteria and method for Korean sentence tokenization. KMOUNLP-NER dataset was processed with the tokenization method on Korean sentences from various resources such as news articles.

Won Ik Cho, et al. cited KMOUNLP-NER on Open Korean Corpora in 2020 [40]. So-Yeop Yoo and Jeong Ok-Ran applied BERT and knowledge graph method for Korean contextual information extraction, trained with KMOUNLP-NER for named entity recognition in 2020 [43]. Yeon-SOO You and Hyuk-Ro Park also used corpuses from this dataset to establish syllable-based Korean NER model [59]. The number of citations of KMOUNLP-NER is underestimated because the author has opened dataset for 2 years without any citation request.

### F. Sci-News-Sum-Kr-50

TABLE XII. SCI-NEWS-SUM-KR-50 OVERVIEW

| Dataset overview | |
|---|---|
| Category | Text Summarization |
| Size | 50 articles |
| Format | JSON |
| License | Noncommercial Research only |
| URL | https://github.com/theeluwin/sci-news-sum-kr-50 |
| Author | Seol, Jinsuk and Lee, Sang-gu |

| Citation text |
|---|
| 설진석, 이상구. "lexrankr: LexRank 기반 한국어 다중 문서 요약." 한국정보과학회 학술발표논문집 (2016): 458-460. |

TABLE XIII. SCI-NEWS-SUM-KR-50 SAMPLE

**Data JSON structure**

```
{
    'title': "Title of the article",
    'source': "Resource URL",
    'slug': "can-be-used-as-unique-id",
    'length': 23,
    'summaries': [0, 1, 4, 7],
    'sentences': [
        "Sentence 1",
        "Sentence 2",
        ...
    ]
}
```

Sci-News-Sum-Kr-50 dataset was proposed by Jinsuk Seol and Sang-gu Lee in 2016 [34]. They collected 50 news articles from IT and Science section of news curation service. Each article is attached with human-written summarization. As copyright of each raw article belongs to the press, this dataset is not available for commercial use. Although the size of this dataset is small, the reliability of this dataset is quite high because the summarized sentences are written by human, without any augmentation algorithm. This dataset was enlisted on Open Korean Corpora [40].

### G. SAE4K

TABLE XIV. SAE4K OVERVIEW

| Dataset overview | |
|---|---|
| Category | Text Summarization |
| Size | 30,837 raw and 50,837 augmented pairs |
| Format | Text |
| License | CC BY-SA 4.0 |
| URL | https://github.com/warnikchow/sae4k |
| Author | Cho, Won Ik, et al. |

| Citation text |
|---|
| Cho, Won Ik, et al. "Discourse component to sentence (DC2S): An efficient human-aided construction of paraphrase and sentence similarity dataset." Proceedings of the 12th Language Resources and Evaluation Conference. 2020. |

TABLE XV. SAE4K SAMPLE

| Sample data pairs (original / summary) | Label |
|---|---|
| 그 외에 다른 물품은 없나요<br># 다른 물품 여부 | 0 |
| 택시를 타는 것이 빠를까 지하철을 타는 것이 빠를까<br># 택시와 지하철 중 빠른 것 | 1 |
| 이번주 전체적인 일기예보가 어떻게 되니<br># 이번주 일기예보 | 2 |
| 후기 작성시 사용하셔도 됩니다. 악평 쓰실거면 사용하지마세요<br># 후기 작성시 악평 쓰지 않기 | 3 |
| 실적이 좋지 않으면 명예퇴직 일 순위라는 걸 명심해<br># 실적 내기 | 4 |
| 총무과에 묻지 말고 관리과에 물어보세요<br># 관리과에 물어보기 | 5 |

SAE4K dataset was published by Won Ik Cho, et al. in 2019 [36]. They opened SAE4K_v1 dataset with 30,837 raw sentence pairs and SAE4K_v2 dataset with 50,837 augmented data. Each data pair has original sentence and extracted structured argument with sentence type label. The labels for sentence categorization and dataset statistics are described on Table XVI. This dataset was enlisted on Open Korean Corpora [40].

TABLE XVI. LABELS OF SAE4K_v1

| Description | Label | Volume | Portion |
|---|---|---|---|
| **Question** | | | |
| Yes / No | 0 | 5,718 | 31.99% |
| Alternative | 1 | 227 | 1.27% |
| Wh- | 2 | 11,924 | 66.73% |
| **Commands** | | | |
| Prohibition | 3 | 477 | 3.67% |
| Requirement | 4 | 12,369 | 95.38% |
| Strong requirement | 5 | 122 | 0.94% |

## H. KLUE

TABLE XVII. KLUE OVERVIEW

**Dataset overview**

| | |
|---|---|
| Category | Language Understanding Evaluation |
| Size | 8 Datasets, total 239k samples |
| Format | - |
| License | CC BY-SA 4.0 |
| URL | https://github.com/KLUE-benchmark/KLUE |
| Author | Park, Sungjoon, et al. |

**Citation text**

Park, Sungjoon, et al. (2021). KLUE: Korean Language Understanding Evaluation. In Thirty-fifth Conference on Neural Information Processing Systems Datasets and Benchmarks Track (Round 2).

TABLE XVIII. KLUE SUBDATASET OVERVIEW

| Name | Type | Size (Train/Dev/Test) |
|---|---|---|
| KLUE-TC (YNAT) | Topic Classification | 45k / 9k / 9k |
| KLUE-STS | Semantic Textual Similarity | 11k / 0.5k / 1k |
| KLUE-NLI | Natural Language Inference | 25k / 3k / 3k |
| KLUE-NER | Named Entity Recognition | 21k / 5k / 5k |
| KLUE-RE | Relation Extraction | 32k / 8k / 8k |
| KLUE-DP | Dependency Parsing | 10k / 2k / 2.5k |
| KLUE-MRC | Machine Reading Comprehension | 12k / 8k / 9k |
| KLUE-DST (WoS) | Dialogue State Tracking | 8k / 1k / 1k |

KLUE is the first Korean-based natural language understanding evaluation dataset. KLUE dataset was created through collaboration among academia and industry, and published by Sungjoon Park, et al. in 2021.

KLUE consists of 8 different large datasets for various NLP tasks. KLUE-TC is a topic classification dataset with 63k data, driven from news headlines. KLUE-STS dataset stands for semantic textual similarity task with 12.5k samples, whose source are news, review and query. KLUE-NLI dataset has 31k samples for natural language inference task, from news, Wikipedia and review. KLUE-NER is a collection of 31k named entity recognition data constructed with news and review articles. KLUE-RE is a relation extraction dataset with 48k datasets driven from Wikipedia and news articles. KLUE-DP dataset consists of 14.5k news and review data for dependency parsing task. KLUE-MRC is a dataset for machine reading comprehension task, with 29k data from Wikipedia and news articles. And the last dataset, KLUE-DST, is a dialogue state tracking dataset with 10k data driven from task oriented dialogue.

As KLUE provides an amount benchmark data for various NLP tasks, it would provide a persuasive criteria for performance measurement on Korean language processing research area.

Myeongjun Jang, et al. has applied KLUE-NLI and KLUE-STS on natural language inference and semantic text similarity experiment for analysis on language understanding models in 2021 [60]. Boseop Kim, et al. has shown the effect and efficiency of large-scale language models by experiments on KLUE-TC and KLUE-STS datasets in 2021 [61].

## I. KorNLU

TABLE XIX. KORNLU OVERVIEW

**Dataset overview**

| | |
|---|---|
| Category | Natural Language Inference, Semantic Textual Similarity |
| Size | 950k data for NLI, 8.6k data for STS |
| Format | TSV |
| License | CC BY-SA 4.0 |
| URL | https://github.com/kakaobrain/KorNLUDatasets |
| Author | Ham, Jiyeon, et al. |

**Citation text**

Ham, Jiyeon, et al. "KorNLI and KorSTS: New Benchmark Datasets for Korean Natural Language Understanding." arXiv preprint arXiv:2004.03289 (2020).

TABLE XX. KORNLI SAMPLE

| KorNLI data example | Label |
|---|---|
| P: 저는, 그냥 알아내려고 거기 있었어요.<br>H: 이해하려고 노력하고 있었어요. | Entailment |
| P: 저는, 그냥 알아내려고 거기 있었어요.<br>H: 나는 처음부터 그것을 잘 이해했다. | Contradiction |
| P: 저는, 그냥 알아내려고 거기 있었어요.<br>H: 나는 돈이 어디로 갔는지 이해하려고 했어요. | Neutral |

TABLE XXI. KORSTS SAMPLE

| KorSTS data example | Label |
|---|---|
| S1. 한 남자가 음식을 먹고 있다.<br>S2. 한 남자가 뭔가를 먹고 있다. | 4.2 |
| S1. 한 비행기가 착륙하고 있다.<br>S2. 애니메이션화된 비행기 하나가 착륙하고 있다. | 2.8 |
| S1. 한 여성이 고기를 요리하고 있다.<br>S2. 한 남자가 말하고 있다. | 0.0 |

KorNLU was established by Jiyeon Ham, et al. in 2020 [14]. It consists of two sub-datasets; KorNLI for natural language inference and KorSTS for semantic textual similarity.

KorNLI has 94k data for train, 2.5k data for development and 5k data for test, driven from SNLI [62], MNLI [63], and XNLI [64]. This dataset classifies the relationship between a pair of sentences into 3 classes: entailment, contradiction, neutral. KorSTS has 5.7k training data, 1.5k development data and 1.4 test data, and the source of KorSTS dataset is STS-B [65]. KorSTS provides a score between 0(dissimilar) and 5(equivalent) for each pair of sentences.

Myeongjun Jang applied KorNLI for experiments on language understanding models in 2021 [60]. Hyunjae Lee proposed a lite BERT model for Korean understanding with benchmark performance experiment on KorNLI and KorSTS

[45]. Yongmin Yoo applied KorSTS for sentence similarity measurement in 2021 [66].

*J. ParaKQC*

TABLE XXII.   ParaKQC Overview

**Dataset overview**

| Category | Semantic Textual Similarity |
|---|---|
| Size | 545k utterances |
| Format | Text |
| License | CC BY-SA 4.0 |
| URL | https://github.com/warnikchow/paraKQC |
| Author | Cho, Won Ik, et al. |

**Citation text**

Cho, Won Ik, et al. "Discourse component to sentence (DC2S): An efficient human-aided construction of paraphrase and sentence similarity dataset." Proceedings of the 12th Language Resources and Evaluation Conference. 2020.

TABLE XXIII.   ParaKQC Sample

**Sentence set examples**

| S1. 메일을 다 비울까 아니면 안읽은 것만 지울까? |
| S2. 메일 중에 안읽은 것만 지울까? 다 지울까? |
| S1. 나한테 집안 습도 조정할 때 어떻게 해야하는 지 좀 알려줄래? |
| S2. 저기요 집안 습도 좀 조정할 수 있는 방법 좀 알려주세요 |
| S3. 오후 말고 밤에는 거실 무드등만 켜놓자 |
| S3. 아침에 말고 밤에 거실 무드등만 켜놓자 |

ParaKQC is a dataset for parallel Korean questions and commands, proposed by Won Ik Cho, et al. in 2020 [36]. It contains 10k utterances that consist with 1,000 sets of 10 similar sentences, provided with an augmentation script which make up whole corpus of 545k utterances.

ParaKQC contributed as a resource of KLUE dataset [24]. Also it is enlisted on Open Korean Corpora [40]. Won Ik Cho, et al. applied the sentence classification scheme of ParaKQC for StyleKQC, a style-variant paraphrase corpus for Korean question and commands [67].

*K. NSMC*

TABLE XXIV.   NSMC Overview

**Dataset overview**

| Category | Semantic Textual Analysis |
|---|---|
| Size | 20k reviews |
| Format | Text |
| License | CC0 1.0 |
| URL | https://github.com/e9t/nsmc |
| Author | Eunjeong Park |

**Citation text**

Eunjeong Park, NSMC: Naver sentiment movie corpus v1.0. (2015). Available at https://github.com/e9t/nsmc

TABLE XXV.   NSMC Sample

| Movie review sample | Label |
|---|---|
| 아 더빙.. 진짜 짜증나네요 목소리 | 0 |
| 흠...포스터보고 초딩영화줄....오버연기조차 가볍지 않구나 | 1 |
| 너무재밓었다그래서보는것을추천한다 | 0 |
| 교도소 이야기구먼 ..솔직히 재미는 없다..평점 조정 | 0 |
| 사이몬페그의 익살스런 연기가 돋보였던 영화!스파이더맨에서 늙어보이기만 했던 커스틴 던스트가 너무나도 이뻐보였다 | 1 |
| 막 걸음마 뗀 3 세부터 초등학교 1 학년생인 8살용영화.ㅋㅋㅋ...별반개도 아까움. | 0 |
| 원작의 긴장감을 제대로 살려내지못했다. | 0 |
| 액션이 없는데도 재미 있는 몇안되는 영화 | 1 |

NSMC dataset was constructed by Eunjeong Park in 2015, with sentences scrapped from Naver movie reviews [37]. Raw materials have user-written comments on the movie with score ratings between 1 (negative) to 10 (positive). The sentences were classified into 3 categories by rating scores; negative (1 ~ 4 score), neutral (5 ~ 8 score), positive (9 ~ 10 score). NSMC only contain the negative (label 0) and positive (label 1) sentences for semantic textual analysis task.

Although NSMC is a highly attractive dataset for Korean NLP, the author has not published related paper or provided official citation request text. Therefore, the exact citation number is not known; search result on Google Scholar indicates that at least 50 papers has applied NSMC for experiment or benchmark test. Yong-Jun Lee, et al. Used NSMC dataset for experiment on Korean-specific emotion annotation model in 2020 [46]. Sangwan Moon and Naoaki Okazaki applied NSMC to evaluate a just-in-time BERT model in 2020 [68]. Kyubong Park, et al. has analyzed the performance benchmark of various Korean NLP with NSMC dataset in 2020 [69]. NSMC also contributed to KLUE [24] and Open Korean Corpora [40].

*L. Toxic Comment*

TABLE XXVI.   Toxic Comment Overview

**Dataset overview**

| Category | Semantic Textual Analysis |
|---|---|
| Size | 20k sentences |
| Format | CSV |
| License | MIT |
| URL | https://github.com/songys/Toxic_comment_data |
| Author | Song, Youngsook |

**Citation text**

Song, Youngsook, Toxic Comment. 2018. Available at https://github.com/songys/Toxic_comment_data.

TABLE XXVII.   Toxic Comment Sample

| Movie review sample | Label |
|---|---|
| 부패한 로마노프 왕조를 기리는 뭣 같은 영화… 온몸으로 항거했던 러시아 민중들이 그저 폭도냐 | 1,0,0,0,0 |
| 빨갱이는 있따… | 1,0,1,0,0 |
| 박찬욱 각본. 박찬욱의 흑역사네. | 1,0,0,1,0 |
| 정일우의 발연기, 깨는 나레이션, 밋밋한 일드풍의 연출.. 화면 때깔만 좋더라! | 1,0,0,0,1 |
| 마이클 더글러스의 여름날 신경질내기 | 1,0,0,1,0 |
| 조폭 살인범 형사놈들 나오는 영화는 예전이나 지금이나 한국영화 단골손님들이구만 ㅉㅉㅉ | 1,0,0,0,1 |
| 그냥 짜증나는 영화, 리암니슨영화는 대체적으로 오바가많은듯 | 1,0,0,1,1 |

Toxic Comment dataset is a collection of negative movie review sentences from NSMC datasets, with more specific emotional labels. The label of each sentence is a 5-dim vector, similar to one-hot expression. The parameters of label vector indicates toxic, obscene, threat, insult, and identity hate. Sentences has at least one '1' indicate for those categories.

*M. KHateSpeech*

TABLE XXVIII. KHATESPEECH OVERVIEW

| Dataset overview | |
|---|---|
| Category | Semantic Textual Analysis |
| Size | Labeled: 9.4k    Unlabeled: 2m |
| Format | TSV |
| License | CC BY-SA 4.0 |
| URL | https://github.com/kocohub/korean-hate-speech |
| Author | Moon, Jihyung, et al. |

**Citation text**

Moon, Jihyung, Won Ik Cho, and Junbum Lee. "BEEP! Korean Corpus of Online News Comments for Toxic Speech Detection." Proceedings of the Eighth International Workshop on Natural Language Processing for Social Media. 2020.

TABLE XXIX. KHATESPEECH SAMPLE

| Comment samples | Gender Bias/ Bias / Hate |
|---|---|
| 송중기 시대극은 믿고본다. 첫회 신선하고 좋았다. | False / None / None |
| 지현우 나쁜놈 | False / None / Offensive |
| 알바쓰고많이만들면되지 돈욕심없으면골 목식당왜나온겨 기댕기게나하고 산에가서 팔어라 | False / None / Hate |
| 설마 ㅈ현정 작가 아니지?? | True / Gender / Hate |

KHateSpeech dataset was published by Jihyung Moon, et al. in 2020 [38]. They scrapped malicious replies from news articles from entertainment and celebrity sections, where the largest amount of negative comments are produced. This dataset provides 9,381 human-labeled data with 2,033,893 unlabeled raw comments. Each comment is annotated on two aspect; existence of social bias and hate speech. And additional binary label has attached to indicate whether a comment contains gender bias or not.

Chanhee Lee, et al. has applied KHateSpeech dataset to analyzed the data efficiency of cross-lingual post-training in pretrained language models in 2021 [70]. Seyoung Lee and Saerom Park used KHateSpeech dataset to proposed a hate speech classification algorithm using ordinal regression in 2021 [47]. Hyeonsang Lee, et al. proposed a CNN based model for toxic comments classification, trained with KHateSpeech dataset in 2020 [71].

*N. 3i4K*

TABLE XXX. 3I4K OVERVIEW

| Dataset overview | |
|---|---|
| Category | Semantic Textual Analysis |
| Size | 19k corpuses, 61k texts, 7k speeches |
| Format | Text |
| License | CC BY-SA 4.0 |
| URL | https://github.com/warnikchow/3i4k |
| Author | Cho, Won Ik, et al. |

**Citation text**

Cho, Won Ik, et al. "Speech intention understanding in a head-final language: A disambiguation utilizing intonation-dependency." arXiv preprint arXiv:1811.04231 (2018).

TABLE XXXI. 3I4K CORPUS COMPOSITION

| Text-based Seiving | Intention | Label | Size |
|---|---|---|---|
| Fragments | - | 0 | 6,538 |
| Clear-cut cases | Statements | 1 | 29,559 |
| | Questions | 2 | 18,418 |
| | Commands | 3 | 17,626 |
| | RQ[a]s | 4 | 3,312 |
| | RC[b]s | 5 | 1,818 |
| Intonation-dependent utterances | unknown | 6 | 5,300 |

[a]Rhetorical Question
[b]Rhetorical Commands

3i4K was published by Won Ik Cho, et al. in 2018 [25]. They collected corpus data from released works of SNU-SLP[1] and Natural Institute of Korean Language[2] to establish FCI (fragments, clear-cut cases, intonation-dependent utterances) dataset. They also manually created questions and commands for dataset.

3i4K dataset is enlisted on Open Korean Corpora [40]. Won Ik Cho, et al. has applied 3i4K to investigate the effectiveness of character-level embedding in Korean sentence classification in 2019 [48].

*O. KAIST Corpus*

KAIST Corpus dataset is a collection of various corpus and machine translation datasets, constructed since 1997 to 2005, published by KAIST Semantic Web Research Center [72]. IT has over 70 million Korean text corpus, position annotated corpus, tree-annotated corpus, Korean-Chinese parallel corpus, Korean-English parallel corpus.

---

[1] http://slp.snu.ac.kr/
[2] https://krean.go.kr

TABLE XXXII. KAIST CORPUS OVERVIEW

**Dataset overview**

| | |
|---|---|
| Category | Various Corpus and Machine Translation |
| Size | 70 million corpuses |
| License | Noncommercial Research Only |
| URL | http://semanticweb.kaist.ac.kr/home/index.php/KAIST_Corpus |
| Provider | Ki-sun Choi |

**Citation text**

Ki-sun Choi (2001), KAIST Language Resources 2001 edition. Result of the core software project from Ministry of Science and Technology, Korea.

TABLE XXXIII. KAIST CORPUS – QUALIFIELD CORPUS

| Dataset Title | Language | Size |
|---|---|---|
| KAIST Raw corpus | Kor | 70m |
| High quality morpho-syntactically annotated corpus | | 1m |
| Automatically Analyzed Large Scale KAIST Corpus | | 40m |
| Korean Tree-Tagging Corpus | | 3k |
| Tree-tagged Corpus 2 | | 30k |
| Chinese-Tagged corpus | | 10k |
| Chinese-English-Korean Multilingual corpus | Kor-Chn-Eng | 60k |
| Chinese-English multilingual corpus | Chn-Eng | 60k |
| Chinese-Korean multilingual corpus | Kor-Chn | 60k |
| Newspaper Corpus (Hankyorh) | Kor | 620 files |
| Newspaper Corpus (Donga-Korean, English, Japanese, Chinese) | Kor-Eng-Jpn-Chn | 1,791 files |

TABLE XXXIV. KAIST CORPUS – PROCESSED RESOURCES

| Dataset Title | Language | Size |
|---|---|---|
| Nouns Definition of General Vocabulary | Kor | 29k |
| Noun Definition Corpus of General Vocabulary | | 29k |
| Co-occurrence data | Kor-Eng | 47m |
| Alignment Model for Extracting English-Korean Translation of Term Constituents | Kor | 24k |
| Terminology corpus– medicine | | 220k |
| Terminology corpus– architectural engineering | | 3.7k |
| Terminology corpus– economics | | 27.7k |
| Terminology corpus– engineering | | 13.6k |
| Terminology corpus– physical metallurgy | | 7.5k |
| Terminology corpus- mechanical engineering | | 50k |
| Terminology corpus– physics | | 107k |
| Terminology corpus– biology | | 84k |
| Terminology corpus– electronic engineering | | 13k |
| Terminology corpus– computer science | | 8.7k |
| Terminology corpus– chemistry | | 66.7k |
| Terminology corpus– chemical engineering | | 20k |
| Terminology corpus– environmental engineering | | 21k |
| Terminology corpus– Adverbs and Frequency | | 2.5k |
| Terminology corpus– Nouns and Frequency | | 4.3m |
| Terminology corpus– Adjectives and Frequency | Kor | 1.7k |
| Terminology corpus– Verbs and Frequency | | 20k |
| Terminology corpus– Verbal case Frames | | 135k |
| Terminology corpus– Co-occurrence | | 16m |
| One-syllable nouns | | 1k |
| Off-line Korean Handwriting Database (Image) | | 1.2k |

KAIST corpus was established by Ki-sun Choi, and published in 2001.[72]. KAIST corpus has various sub-datasets described on Table XXXIII and TABLE XXXIV.

Qualified corpus has 7 Korean corpus datasets and 4 multilingual corpus datasets. Processed resources section provides 23 Korean corpus dataset and 1 Korean-English co-occurrence dataset, and an off-line Korean handwriting image dataset. KAIST corpus is also enlisted on Open Korean Corpora [40].

Considering the year of publication, the number of citations of KAIS corpus dataset is very low. So it's difficult to find a benchmark performance of recent algorithms tested with KAIST corpus. However, as the size of datasets are very large and the quality of each dataset is excellent, it still worth using for algorithm design and model training.

## IV. DISCUSSION

As the interests and demands of companies on Korean language processing dataset has dramatically increased, fund and investment seems to be increased too. Therefore, great dataset such as KLUE has been published and accepted to Neural IPS this year. As described on Table 1, most of the datasets popular today were published within last 3 years. This reflects the recent advance in popularity of Korean NLP dataset construction.

Since the datasets were recently published, I guess that many research projects using them are currently in progress, or have not yet published. This may explain the low citation scores of useful datasets. The number of citation is considered as a good indicator to evaluate a popularity of dataset. I agree that a dataset which are frequently cited is very useful to produce a reasonable benchmark result. However, many dataset papers are cited because of the novel method proposed by the authors, rather than the dataset itself.

A researcher should not underestimate or overestimate a dataset by the paper's citation score only. I suggest exploring various datasets to find one, best appropriate for your research purpose.

## V. Conclusion

This paper provided a survey of Korean-based natural language processing datasets. The survey section proposed brief and comprehensive summarization and detailed information. It covered conversation log, question answering, named entity recognition, text summarization, natural language understanding, semantic sentence similarity, semantic textual analysis, semantic speech analysis and machine translation. The discussion outlined a brief guide to consider when choosing a dataset. In conclusion, I hope this paper to help researchers' to reduce time consuming works for dataset exploring and to choose appropriate dataset in Korean NLP area.

## Acknowledgment

This paper was inspired while writing the book *"142 datasets for AI researchers."* with Life & Power Press Co., Ltd. I would like to express my gratitude to Jehoon Yoo for suggesting me collect various datasets to write such a book.

## References

[1] Palangi, Hamid, et al. "Deep sentence embedding using long short-term memory networks: Analysis and application to information retrieval." *IEEE/ACM Transactions on Audio, Speech, and Language Processing* 24.4 (2016): 694-707.

[2] Luukkonen, Petri, Markus Koskela, and Patrik Floréen. "LSTM-based predictions for proactive information retrieval." *arXiv preprint arXiv:1606.06137* (2016).

[3] Palangi, Hamid, et al. "Semantic modelling with long-short-term memory for information retrieval." *arXiv preprint arXiv:1412.6629* (2014).

[4] Dauphin, Yann N., et al. "Language modeling with gated convolutional networks." *International conference on machine learning*. PMLR, 2017.

[5] Yang, Xi, et al. "Measurement of Semantic Textual Similarity in Clinical Texts: Comparison of Transformer-Based Models." JMIR medical informatics 8.11 (2020): e19735.

[6] Sun, Lichao, et al. "Mixup-Transformer: Dynamic Data Augmentation for NLP Tasks." *arXiv preprint arXiv:2010.02394* (2020).

[7] Wolf, Thomas, et al. "Huggingface's transformers: State-of-the-art natural language processing." *arXiv preprint arXiv:1910.03771* (2019).

[8] Gillioz, Anthony, et al. "Overview of the Transformer-based Models for NLP Tasks." *2020 15th Conference on Computer Science and Information Systems (FedCSIS)*. IEEE, 2020.

[9] Zhuang, Hang, et al. "Natural language processing service based on stroke-level convolutional networks for Chinese text classification." *2017 IEEE international conference on web services (ICWS)*. IEEE, 2017.

[10] Cui, Yiming, et al. "Revisiting pre-trained models for chinese natural language processing." *arXiv preprint arXiv:2004.13922* (2020).

[11] Al-Smadi, Mohammad, et al. "Transfer learning for Arabic named entity recognition with deep neural networks." *Ieee Access* 8 (2020): 37736-37745.

[12] Romeo, Salvatore, et al. "Language processing and learning models for community question answering in arabic." Information Processing & Management 56.2 (2019): 274-290.

[13] Yun, Hyungbin, Ghudae Sim, and Junhee Seok. "Stock prices prediction using the title of newspaper articles with korean natural language processing." *2019 International Conference on Artificial Intelligence in Information and Communication (ICAIIC)*. IEEE, 2019.

[14] Ham, Jiyeon, et al. "KorNLI and KorSTS: New Benchmark Datasets for Korean Natural Language Understanding." *Proceedings of the 2020 Conference on Empirical Methods in Natural Language Processing: Findings*. 2020.

[15] Stratos, Karl. "A Sub-Character Architecture for Korean Language Processing." Proceedings of the 2017 Conference on Empirical Methods in Natural Language Processing. 2017.

[16] Hwang, Myeong-Ha, et al. "KoRASA: Pipeline Optimization for Open-Source Korean Natural Language Understanding Framework Based on Deep Learning." *Mobile Information Systems* 2021 (2021).

[17] Jung, Haemin, and Wooju Kim. "Automated conversion from natural language query to SPARQL query." *Journal of Intelligent Information Systems* (2020): 1-20.

[18] Kim, Hayan, et al. "Visual language approach to representing KBimCode-based Korea building code sentences for automated rule checking." *Journal of Computational Design and Engineering* 6.2 (2019): 143-148.

[19] J.-H. Kim and G. C. Kim, Guideline on Building a Korean Part-of-Speech Tagged Corpus: KAIST Corpus, Technical Report CS-TR-95-99, Department of Computer Science, KAIST, 1995 (in Korean).

[20] 천민아, 다중 생성 단위의 관계 점수를 이용한 학습 말뭉치 생성: 개체명 말뭉치를 중심으로, 한국해양대교 대학원 컴퓨터공학과, 박사학위논문: 134, 2021 년 2 월.

[21] Song, Youngsook, Toxic Comment. 2018. Available at https://github.com/songys/Toxic_comment_data.

[22] Alan Kang and Jieun Kim, Petition. 2017. Available at https://github.com/akngs/petitions.

[23] Ha, Jung-Woo, et al. "ClovaCall: Korean goal-oriented dialog speech corpus for automatic speech recognition of contact centers." arXiv preprint arXiv:2004.09367 (2020).

[24] Park, Sungjoon, et al. (2021). KLUE: Korean Language Understanding Evaluation. In Thirty-fifth Conference on Neural Information Processing Systems Datasets and Benchmarks Track (Round 2).

[25] Cho, Won Ik, et al. "Speech intention understanding in a head-final language: A disambiguation utilizing intonation-dependency." arXiv preprint arXiv:1811.04231 (2018).

[26] Zeng, Zhiqiang, et al. "Survey of natural language processing techniques in bioinformatics." *Computational and mathematical methods in medicine* 2015 (2015).

[27] Otter, Daniel W., Julian R. Medina, and Jugal K. Kalita. "A survey of the usages of deep learning for natural language processing." IEEE Transactions on Neural Networks and Learning Systems 32.2 (2020): 604-624.

[28] Oshikawa, Ray, Jing Qian, and William Yang Wang. "A survey on natural language processing for fake news detection." *arXiv preprint arXiv:1811.00770* (2018).

[29] Han, Manhui, et al. "Natural language processing on Korean language: A survey." *Processing of the Korean Information Science Society Conference.* (2016): pp 681-683.

[30] Song, Yeongsook. "Awesome Korean Data." Available on GitHub, at https://github.com/songys/AwesomeKorean_Data.

[31] Song, Youngsook, Chatbot_data. 2018. Available on GitHub at https://github.com/songys/Chatbot_data.

[32] Lim, Seungyoung, Myungji Kim, and Jooyoul Lee. "KorQuAD1. 0: Korean QA dataset for machine reading comprehension." arXiv preprint arXiv:1909.07005 (2019).

[33] 박혜웅, 송영숙. "음절 기반의 CNN 을 이용한 개체명 인식." 한국어정보학회 학술대회 (2017): 330–332.

[34] 설진석, 이상구. "lexrankr: LexRank 기반 한국어 다중 문서 요약." 한국정보과학회 학술발표논문집 (2016): 458–460.

[35] Cho, Won Ik, et al. "Machines getting with the program: Understanding intent arguments of non-canonical directives." arXiv preprint arXiv:1912.00342 (2019).

[36] Cho, Won Ik, et al. "Discourse component to sentence (DC2S): An efficient human-aided construction of paraphrase and sentence similarity dataset." Proceedings of The 12th Language Resources and Evaluation Conference. 2020.

[37] Eunjeong Park, NSMC: Naver sentiment movie corpus v1.0. (2015). Available at https://github.com/e9t/nsmc

[38] Moon, Jihyung, Won Ik Cho, and Junbum Lee. "BEEP! Korean Corpus of Online News Comments for Toxic Speech Detection." Proceedings of the Eighth International Workshop on Natural Language Processing for Social Media. 2020.

[39] Yun, Hyeonseo, and Sunyong Yoo. "Transformer-based Unethical Sentence Detection." Journal of Digital Contents Society 22.8 (2021): 1289-1293.

[40] Cho, Won Ik, Sangwhan Moon, and Youngsook Song. "Open Korean Corpora: A Practical Report." arXiv preprint arXiv:2012.15621 (2020).

[41] Lee, Hanbum, Jahwan Koo, and Ung-Mo Kim. "A Study on Emotion Analysis on Sentence using BERT." Proceedings of the Korea Information Processing Society Conference. Korea Information Processing Society, 2020.

[42] Hyunjoong Kim et al. Korpora:Korean Corpora Archives. 2019. Available at https://github.com/ko-nlp/Korpora

[43] Yoo, SoYeop, and OkRan Jeong. "Korean Contextual Information Extraction System using BERT and Knowledge Graph." Journal of Internet Computing and Services 21.3 (2020): 123-131.

[44] Kim, Boseop, et al. "What Changes Can Large-scale Language Models Bring? Intensive Study on HyperCLOVA: Billions-scale Korean Generative Pretrained Transformers." arXiv preprint arXiv:2109.04650 (2021).

[45] Lee, Hyunjae, et al. "KoreALBERT: Pretraining a Lite BERT Model for Korean Language Understanding." 2020 25th International Conference on Pattern Recognition (ICPR). IEEE, 2021.

[46] Lee, Young-Jun, Chae-Gyun Lim, and Ho-Jin Choi. "Korean-Specific Emotion Annotation Procedure Using N-Gram-Based Distant Supervision and Korean-Specific-Feature-Based Distant Supervision." Proceedings of The 12th Language Resources and Evaluation Conference. 2020.

[47] Lee, Seyoung, and Saerom Park. "Hate Speech Classification Using Ordinal Regression." Proceedings of the Korean Society of Computer Information Conference. Korean Society of Computer Information, 2021.

[48] Cho, Won Ik, Seok Min Kim, and Nam Soo Kim. "Investigating an effective character-level embedding in korean sentence classification." arXiv preprint arXiv:1905.13656 (2019).

[49] Ruffieux, Simon, et al. "A survey of datasets for human gesture recognition." International conference on human-computer interaction. Springer, Cham, 2014.

[50] Bang, Jihwan, et al. "Boosting Active Learning for Speech Recognition with Noisy Pseudo-labeled Samples." arXiv preprint arXiv:2006.11021 (2020).

[51] Cho, Won Ik, et al. "Kosp2e: Korean Speech to English Translation Corpus." arXiv preprint arXiv:2107.02875 (2021).

[52] Bang, Jeong-Uk, et al. "Ksponspeech: Korean spontaneous speech corpus for automatic speech recognition." Applied Sciences 10.19 (2020): 6936.

[53] Lee, Sang-Woo, et al. "Carecall: a call-based active monitoring dialog agent for managing covid-19 pandemic." arXiv preprint arXiv:2007.02642 (2020).

[54] Hwang, Sewoong, and Jonghyuk Kim. "Toward a Chatbot for Financial Sustainability." Sustainability 13.6 (2021): 3173.

[55] Rajpurkar, Pranav, Robin Jia, and Percy Liang. "Know what you don't know: Unanswerable questions for SQuAD." arXiv preprint arXiv:1806.03822 (2018).

[56] 이동헌, et al. "BERT 를 이용한 한국어 기계 독해." 한국정보과학회 학술발표논문집 (2019): 557-559.

[57] Kim, Minho, et al. "Machine Reading Comprehension-based Question and Answering System for Search and Analysis of Safety Standards." Journal of Korea Multimedia Society 23.2 (2020): 351-360.

[58] 정서형, and 곽노준. "BERT 를 이용한 한국어 질의응답 데이터셋에서의 기계 독해." 대한전자공학회 학술대회 (2020): 625-630.

[59] You, Yeon-Soo, and Hyuk-Ro Park. "Syllable-based Korean named entity recognition using convolutional neural network." 한국마린엔지니어링학회지 44.1 (2020): 68-74.

[60] Jang, Myeongjun, Deuk Sin Kwon, and Thomas Lukasiewicz. "Accurate, yet inconsistent? Consistency Analysis on Language Understanding Models." arXiv preprint arXiv:2108.06665 (2021).

[61] Kim, Boseop, et al. "What changes can large-scale language models bring? intensive study on hyperclova: Billions-scale korean generative pretrained transformers." arXiv preprint arXiv:2109.04650 (2021).

[62] Bowman, Samuel R., et al. "A large annotated corpus for learning natural language inference." Conference on Empirical Methods in Natural Language Processing, EMNLP 2015. Association for Computational Linguistics (ACL), 2015.

[63] Williams, Adina, Nikita Nangia, and Samuel R. Bowman. "A Broad-Coverage Challenge Corpus for Sentence Understanding through Inference." NAACL-HLT. 2018.

[64] Conneau, Alexis, et al. "XNLI: Evaluating cross-lingual sentence representations." 2018 Conference on Empirical Methods in Natural Language Processing, EMNLP 2018. Association for Computational Linguistics, 2020.

[65] Cer, Daniel, et al. "SemEval-2017 Task 1: Semantic Textual Similarity Multilingual and Crosslingual Focused Evaluation." Proceedings of the 11th International Workshop on Semantic Evaluation (SemEval-2017). 2017.

[66] Yoo, Yongmin, et al. "A novel hybrid methodology of measuring sentence similarity." Symmetry 13.8 (2021): 1442.

[67] Cho, Won Ik, et al. "StyleKQC: A Style-Variant Paraphrase Corpus for Korean Questions and Commands." arXiv preprint arXiv:2103.13439 (2021).

[68] Moon, Sangwhan, and Naoaki Okazaki. "PatchBERT: Just-in-Time, Out-of-Vocabulary Patching." Proceedings of the 2020 Conference on Empirical Methods in Natural Language Processing (EMNLP). 2020.

[69] Park, Kyubyong, et al. "An Empirical Study of Tokenization Strategies for Various Korean NLP Tasks." arXiv preprint arXiv:2010.02534 (2020).

[70] Lee, Chanhee, et al. "Exploring the Data Efficiency of Cross-Lingual Post-Training in Pretrained Language Models." Applied Sciences 11.5 (2021): 1974.

[71] 이현상, 이희준, and 오세환. "딥러닝 기술을 활용한 악성댓글 분류: Highway Network 기반 CNN 모델링 연구." 한국경영학회 통합학술발표논문집 (2020): 343-351.

[72] Ki-sun Choi (2001), KAIST Language Resources 2001 edition. Result of the core software project from Ministry of Science and Technology, Korea.